\documentclass[a4]{article} 
\usepackage{times}  
\usepackage{helvet}  
\usepackage{courier}  
\usepackage[hyphens]{url}  
\usepackage{graphicx} 
\usepackage{natbib}  
\usepackage{caption} 
\usepackage{algorithm}
\usepackage{algorithmic}

\usepackage[margin=2cm]{geometry}

\usepackage{newfloat}
\usepackage{listings}
\DeclareCaptionStyle{ruled}{labelfont=normalfont,labelsep=colon,strut=off} 
\floatstyle{ruled}
\newfloat{listing}{tb}{lst}{}
\floatname{listing}{Listing}
\title{Relevance Score: A Landmark-Like Heuristic for Planning}
\author {
    Oliver Kim\textsuperscript{1,2},
    Mohan Sridharan\textsuperscript{1,3}\\
    \small
    \textsuperscript{1} University of Birmingham, \textsuperscript{2} University of Melbourne, \textsuperscript{3} University of Edinburgh
}

\usepackage{amsmath}
\usepackage{amssymb}
\usepackage{mathtools}

\usepackage{caption}
\usepackage{subcaption}
\usepackage{multirow}

\begin{document}

\maketitle

\begin{abstract}
Landmarks are facts or actions that appear in all valid solutions of a planning problem. They have been used successfully to calculate heuristics that guide the search for a plan. We investigate an extension to this concept by defining a novel "relevance score" that helps identify facts or actions that appear in most but not all plans to achieve any given goal. We describe an approach to compute this relevance score and use it as a heuristic in the search for a plan. We experimentally compare the performance of our approach with that of a state of the art landmark-based heuristic planning approach using benchmark planning problems. While the original landmark-based heuristic leads to better performance on problems with well-defined landmarks, our approach substantially improves performance on problems that lack non-trivial landmarks. 

\end{abstract}

\section{Introduction}

\label{sec:intro}
The use of heuristics to guide search and limit the search space is an important component of modern planning systems. There is a well-established literature of methods that use heuristics to improve the computational efficiency of computing plans. The existence of a sound heuristic that can be computed quickly makes path planning in Euclidean space much more efficient than the more abstract search spaces used for task planning. One of the successful heuristics used for task planning is a count of the "\textit{landmarks}" that remain to be reached from a given state~\cite{Zhu2003,Hoffmann2004,Richter2010,Keyder2010}, where a landmark is a fact, action, or a logical formula over facts and/or actions, that is present in all valid solutions (i.e., sequence of actions) for a planning problem. This leads to a preference for actions (in plans) that are a landmark or achieve one. However, the task of computing all landmarks for a planning problem is expensive; it is known to be PSPACE-complete~\cite{Porteous2001a}. Systems that compute landmarks thus identify a subset of key landmarks, often using a relaxed version of the planning problem~\cite{Keyder2010}, but the cost of computing landmarks is still significant.

A common limitation of most approaches that compute and use landmarks to guide planning is that such landmarks must exist for the planning problems under consideration.
In many complex domains where there are multiple possible routes to the goal, the only simple (single fact) landmarks that exist may be trivial, i.e., those present in the goal specification or already true in the initial state. While complex logical formulae may serve as landmarks, it can be challenging to compute and use such logical formulae in a heuristic. The approach described in this paper avoids this limitation by computing a \textit{relevance score}, which also considers the value of achieving facts (or actions) that appear in many but not all valid plans. The underlying hypothesis is that this additional information will lead to a more efficient search for a valid solution to the given planning problem. As the baseline planner, we use LAMA~\cite{Richter2010}, which uses a combination of landmark counting and the Fast Forward system~\cite{Hoffmann2001}, with the shortest plan found using weighted $A^*$ search under delete relaxation conditions. It has been considered state of the art at planning competitions for over a decade. 

\begin{figure}[tb]
\begin{subfigure}{\linewidth}
    \includegraphics[width=\linewidth]{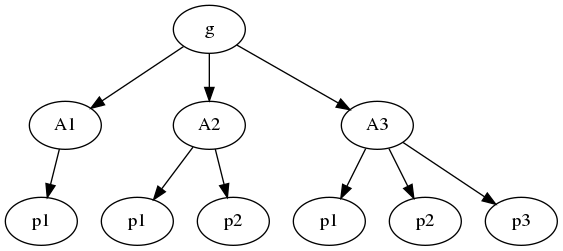}
\end{subfigure}
\caption{\textbf{Illustration of the distinction between landmarks and relevant facts} \textit{p1} is a landmark because all partial plans that achieve the goal~\textit{g} must use one of actions \textit{A1}, \textit{A2}, or \textit{A3}, for which it is a precondition. \textit{p2} is not a landmark because the goal could be achieved by \textit{A1}, but it is highly \textit{relevant} because there is a $\frac{2}{3}$ chance of it being present in any partial plan to achieve the goal. }
\label{fig:illustration}
\end{figure}

The \textbf{key contribution} of this paper is to define and describe an approach for computing a novel "\textit{Relevance Score}" heuristic ($h_{\Xi})$. This heuristic that generalizes the concept of a landmark, which must be true at some point in \emph{all} plans (under delete relaxation), to relevance, which evaluates facts or actions according to how often they appear in plans (also under the delete relaxation). Figure~\ref{fig:illustration} illustrates the information we aim to capture with the relevance score, and how it relates to landmarks. For ease of comparison with the baseline LAMA planner, we limit our focus to facts as the \textit{landmark-like} entities. We then demonstrate through experimental evaluation involving individual and combinations of benchmark planning problems that while the original landmark-based heuristic leads to better performance on problems with
well-defined landmarks, our approach based on relevance scores substantially improves performance on problems that lack non-trivial landmarks.

The remainder of this paper is organized as follows. Section~\ref{sec:rel-work} motivates our proposed approach by discussing related work. Section~\ref{sec:definitions} describes our problem formulation and the proposed heuristic, while Section~\ref{sec:methods} describes the calculation of this heuristic and its use to compute plans. Experimental results described in Section~\ref{sec:results} form the basis of the discussion of next steps in Section~\ref{sec:discuss}.

\section{Related work}
\label{sec:rel-work}
Landmarks as a heuristic were initially described in~\cite{Porteous2001a} and developed further in~\cite{Hoffmann2004}, as an extension to goal ordering. They sought to find and order facts that had to be achieved in any valid plan.  This partial ordering of landmarks was treated as a sequence of sub-goals to break a planning problem into sub-problems. Landmarks are usually found using a relaxed planning graph (RPG)~\cite{Hoffmann2001}, which performs reachability analysis under delete relaxation conditions and requires multiple steps to check that landmarks and their order are sound.

Many researchers have used landmarks to guide heuristic search planners~\cite{Zhu2003,Helmert2006,Richter2008}. Others extended this idea to find cost-optimal plans and began to use actions and propositional formulas of facts as potential landmarks~\cite{Keyder2008,Karpas2009,Richter2010}. Most use a landmark counting heuristic, which assumes that each landmark that has yet to be reached will cost one action. The estimate of distance to the the goal is then given by the number of landmarks remaining to be found. There has been some work on heuristics that preserve action costs, and consider actions that can simultaneously achieve multiple landmarks~\cite{Helmert2009,Karpas2009}. These and other papers explored more efficient ways to identify many landmarks~\cite{Keyder2010}.

Researchers have explored different ways to compute and use landmarks. One example is the translation of Relaxed Planning Graphs (RPGs) into AND/OR graphs for which landmarks are unique maximal solutions computed using Bellman-Ford methods~\cite{Keyder2010}. Applying delete relaxation to a planning problem involves a loss of information, and can lead to actions or action sequences that are impossible in the original problem. One approach compiles away negative preconditions and effects, while preserving the information they contained~\cite{Haslum2009}. By considering pairs ($m=2$), triplets ($m=3$) or larger conjunctions of facts, the requirement to achieve them simultaneously is preserved, resulting in a larger problem $ \Pi^m $ free from negative preconditions or effects. Others have built on this idea~\cite{Keyder2010}. Since the size of the problem grows exponentially with $m$, and introducing new, compound facts into the domain poses other challenges, methods that use landmarks often prefer to just accept the limitations of delete relaxation. 

There continues to be work on computing and using landmarks. This includes computing landmarks using ungrounded or "lifted" planning domain definitions, which allows some of the complexity of disjunctive landmarks to be represented in the range of possible groundings~\cite{Wichlacz2022}. It also includes the use of ordering information to calculate a new heuristic, which improves its accuracy~\cite{Buchner2021}. Landmark-based heuristics are also used in many related fields such as goal recognition~\cite{Pereira2017,Vered2018} and contingent planning~\cite{Maliah2018,Segovia-Aguas2022}.  

\section{Definitions and Background Knowledge}
\label{sec:definitions}
We begin by defining the notation and some concepts used in this paper, followed by the definition of our proposed relevance score heuristic.

\subsection{Classical planning problem}
A classical planning problem is defined by the tuple~\cite{GNT2}: 
\begin{equation*}
    \Pi = \langle F, A, I, G\rangle
\end{equation*}
where $F$ is a finite set of ground predicates representing facts; $A$ is a finite set of actions; $I\subseteq F$ is the set of facts true in the initial state; and $G \subseteq F$ is the set of facts representing goals. A \emph{state} $\sigma$ is a set of facts that are true at a particular time. An action $a \in A$ has preconditions $pre(a)$ and effects $ef\!f(a)$, each of which is subdivided into sets of positive and negative facts: $ pre^+(a), pre^-(a), ef\!f^+(a),  ef\!f^-(a) $. If an action's positive (negative) preconditions are true (false) in a particular state, it is applicable in that state:
\begin{equation*}
    Applicable(a, \sigma) \iff \sigma \models pre^+(a)\cap\neg pre^-(a)
\end{equation*}
When an action is applied to a state in which it is applicable, the resultant state is given by:
\begin{equation*}
    Result(a, \sigma)  \models \sigma \cup ef\!f^+(a) \setminus ef\!f^-(a)
\end{equation*}
A sequence of actions $[a_1 ... a_n]$ is applicable in a state if each action is applicable in the state resulting from the previous sequence of actions, i.e., $ Applicable(a_1, \sigma), Applicable(a_2, Result(a_1, \sigma))$ etc. The result of applying a sequence of actions to a state in which it is applicable is the result of applying each action in that sequence, i.e., $ Result([a_1 ... a_n], \sigma)  = Result(a_n, Result(a_{n-1}, ...Result(a_1, \sigma))$.

A plan$ \pi^\Pi$ for problem $ \Pi = \langle F, A, I, G\rangle$ is a sequence of actions $[a_1 ... a_n]$ that is applicable in $I$ and results in a state where all facts in $G$ are true, i.e., $Result(\pi^\Pi, I) \models G$.

\subsection{Landmarks}
A landmark is defined as a propositional formula of facts that is true at some point in the execution of all valid plans. Due to the difficulty in identifying disjunctive landmarks, many planning systems including LAMA~\cite{Richter2010} only make use of fact landmarks. These are single facts that must be true at some point in the execution of all valid plans. Each fact in the goal and the initial state is a landmark, but is of little use to a planner; these are referred to as \textit{trivial landmarks}. Facts that must be true at some point other than at the start or end of a plan execution are considered \textit{non-trivial}.

The relevance score that we propose in this paper applicable to both fact-based and action-based landmark-like heuristics. However, to facilitate use and comparison with LAMA, we limit our discussion to facts in this paper.

\subsection{Simplifying approximations}
A classical planning problem can be simplified for computing heuristics that will assist in solving the original problem~\cite{Bonet2001,Hoffmann2001}. One such simplification is  \textit{delete relaxation}, which removes all negative preconditions and negative effects from all actions. Another simplification is often used for planning domains with multiple goals. Specifically, 
the domain is modified by adding action $achieveGoal$ that has a single positive effect $ef\!f^+(achieveGoal) = \{success\}$, and the problem's goals as preconditions $pre^+(achieveGoal) = G$. Then, $success$ is used as the goal for computing heuristics, allowing all goals to be considered jointly.

\subsection{Backtracking tree}
A tree is a standard representation of a domain (and changes in the domain) for a planning problem. A tree consists of nodes $\underline{n}$ and edges $e$. A node $\underline{n} = \langle l, E \rangle$ consists of a label $ label(\underline{n}) = l \in F\cup A$, which references a fact $f$ or action $a$, and a set of edges $E$. An edge $ e = (\underline{n} \rightarrow \underline{c})$ links a parent node $\underline{n}$ to a child node $\underline{c}$. 

Given an edge $e = (\underline{n} \rightarrow \underline{c})$, the function $parent(\underline{c})$ yields $\underline{n}$. The function $children(\underline{n})$ yields a set of nodes such that for each node $c$, $ parent(\underline{c}) = \underline{n}$. The root of a tree is the only node with no parent, i.e., $parent(\underline{r}) = None$. 

A delete-relaxed planning problem $\Pi = \langle F, A, I, G\rangle$ defines a tree $T_\Pi$. Node $\underline{r}$ is the root of $T_\Pi$ and has $label (\underline{r}) = achieveGoal$. An action node $\underline{a}$, i.e., a node whose label is an action, has children whose labels are its preconditions $f \in pre(label(\underline{a}))$. A fact node $\underline{f}$ has children whose labels are actions $a$ such that $label(\underline{f}) \in ef\!f(a)$. 

The function $L(l)$ maps a label $l$ to a tree's nodes that have that label: 
\begin{align} \nonumber
    L(l) &= 
        \{ \forall \underline{n} : label(\underline{n}) = l\}
    \\
\intertext{%
    The path of a node is the sequence of alternating fact and action nodes that is generated by adding the parent of the current node until the root is reached: 
}
\nonumber
    T_{\Pi}^{path}(\underline{n}) &= 
        [\underline{n}, parent(\underline{n}), ..., \underline{r}] 
    \\
\intertext{%
    The descendants of a node $ \underline{n} $ are all nodes that have $ \underline{n} $ in their path:
}
\nonumber
    T_{\Pi}^{desc}(\underline{n}) &= 
        \{\forall \underline{d}: \underline{n} \in T_{\Pi}^{path}(\underline{d}) \}
\end{align}
Actions are excluded from the children of a fact node $ \underline{f} $ if any preconditions of that action appear as labels on any node in $ path(\underline{f}) $. This prevents cycles, which would represent unachievable requirements.

\subsection{Non-deterministic agent}
The aim of the \textit{relevance score} heuristic is to estimate how frequently a fact must become true in some distribution of partial plans. We use the behaviour of a hypothetical Non-deterministic agent (NDA) do define this distribution. A sub-tree~$ S_{\Pi} $ of $ T_{\Pi} $, consists of some subset of nodes in $ T_{\Pi} $ and their paths, chosen according to Algorithm~\ref{alg:subtree}. Applying the sequence of actions represented by the path of each leaf-node within $ S_{\Pi} $ will result in the satisfaction of the goal. Thus, $ S_{\Pi} $ may be considered a partial plan under delete-relaxation conditions.

Our interpretation, which we pursue in this paper, is that a high probability of being sampled by such an NDA indicates that a fact is highly relevant to achieving the goal. Facts that appear in all partial plans that could be sampled must be in all plans, and so are landmarks.



\begin{algorithm}[tb]
\begin{algorithmic}[1]
\STATE Let $ S_{\Pi} \leftarrow \{\underline{root}\} $ 
\STATE Let $ frontier $ be a stack
\STATE $ frontier.push(\underline{root}) $
\WHILE{$ frontier $ is not empty }
\label{algline:whilefrontstart}
    \STATE $ \underline{n} \leftarrow frontier.pop()$ 
    \IF{$ label(\underline{n}) \in F $ }
    \label{algline:factstart}
        \STATE Let $ \underline{m} \leftarrow choose(children(\underline{n})) $
        \STATE $ frontier.push(\underline{m}) $ 
        \STATE Add $\underline{m}$ to $S_{\Pi}$ 
    \label{algline:factend}
    \ELSE
    \label{algline:actstart}
        \FOR{$ \underline{m} \in children( \underline{n} )$ }
            \STATE $ frontier.push(\underline{m}) $ 
            \STATE Add $\underline{m}$ to $S_{\Pi}$ 
        \ENDFOR
    \label{algline:actend}
    \ENDIF
    \STATE Remove $ \underline{n} $ from frontier
\label{algline:whilefrontend}
\ENDWHILE
\end{algorithmic}
\caption{\textbf{An NDA sampling sub-tree $ S_{\Pi} $ from $ T_{\Pi} $.} 
Lines~\ref{algline:whilefrontstart}~-~\ref{algline:whilefrontend} add required sub-goals and actions that require them until either a sub-goal is TRUE in the state, or there is no action available that could achieve it. 
Lines~\ref{algline:factstart}~-~\ref{algline:factend} choose one action that achieves the required fact. 
Lines~\ref{algline:actstart}~-~\ref{algline:actend} require all preconditions of an action.
}
\label{alg:subtree}
\end{algorithm}

\subsection{Relevance score}

Let the \emph{relevance score}~$ \Xi(l) $ represent the probability that a node with label $ l $ would be sampled by the NDA:
\begin{align}
\nonumber
    \Xi(l) &= 
        P \left( \exists \underline{l} \in S_{\Pi} | label(\underline{l}) = l \right)
\end{align}
Let the \emph{local relevance score}~$ \Xi(l, \underline{n}) $ represent the probability that a node with label $ l $ would be sampled by the NDA, given that node $ \underline{n} $  (and thus $ S_{\Pi}^{path}(\underline{n}) $) has been sampled. 
\begin{align}
\label{eqn:localrel}
    \Xi(l, \underline{n}) &=                                           
        P \left( \exists \underline{l} \in S_{\Pi}^{desc}(\underline{n}) | label(\underline{l}) = l \right) 
\end{align}
If $ label(\underline{n}) $ is a fact, any one of its children could be sampled, and may have a node with label $ l $ among its descendants:
\begin{align} 
\nonumber
    \shortintertext{$ \texttt{if } label(n) \in F: $ }
\nonumber
    \Xi(l, \underline{n}) &=                                            
        \sum_{\underline{c} \in children(\underline{n})} 
        P \left( \underline{c} \in S_{\Pi}^{desc}(\underline{n}) \right) 
        \times 
        \Xi(l, \underline{c}) 
\end{align}
The NDA chooses one child of a fact node with uniform probability, which implies:
\begin{align} 
\label{eqn:XiFactBasic}
    \Xi(l, \underline{n}) &=
        \sum_{\underline{c} \in children(\underline{n})} 
        \frac
        {
            \Xi(l, \underline{c})
        }
        {
            \left| children(\underline{n}) \right|
        }
\end{align}
If $ label(\underline{n}) $ is an action, then all its children will be sampled. $\Xi(l, \underline{n})$ is thus 1 minus the probability that none of their sampled descendants have label $l$:
\begin{align} 
\nonumber
    \shortintertext{$ \texttt{if } label(n) \in A:   $ }
\label{eqn:XiActBasic}
    \Xi(l, \underline{n}) &= 
        1 - \prod_{\underline{c} \in children(\underline{n})} 
        \left( 
            1 - \Xi(l, \underline{c})
        \right)
\end{align}

\subsection{Choices counter}
The probability of a node $\underline{n}$ being sampled by the NDA, $\xi(\underline{n}) = P(\underline{n} \in S_{\Pi})$ depends on the number of alternative action choices that could have been made instead of those that reach that node. The tree's root will always be sampled:
\begin{align}
\nonumber
    \xi(\underline{r}) &= 1
\end{align}
Between an action and its preconditions, the NDA has no choices to make, so $ \xi $ is passed down unchanged:
\begin{align}
\nonumber
    \xi(\underline{f}) &= \xi(parent(\underline{f}))
\end{align}
The NDA chooses one action that could supply a fact from the set of actions that make up $children(f)$:
\begin{align}
\label{eqn:CCAct}
    \xi(\underline{a}) &= \frac{\xi(parent(\underline{a}))}{\left| children(parent(\underline{a})) \right|}
\end{align}
where $\underline{r}$, $\underline{f}$, $\underline{a}$ are root, fact, and action nodes respectively.

\subsection{Lowest common ancestors}
The Lowest Common Ancestor (LCA) of two nodes $\underline{n}$ and $\underline{m}$ is the lowest node in the tree which is an ancestor of both nodes:
\begin{equation}
\nonumber
    LCA(\underline{n},\underline{m}) = \underset{\underline{i} \in T^{path}_{\Pi}(\underline{n}) \cap T^{path}_{\Pi}(\underline{m})}{argmax}(|T^{path}_{\Pi}(\underline{i})|)
\end{equation}
This computation is associative and can generalize to any number of nodes.

\section{Calculating the Relevance Score}
\label{sec:methods}
Next, we describe our approach to compute the proposed relevance score heuristic. We begin with some basics.
The code used to implement and test this will be made available~\footnote{\url{https://github.com/relevancescoreplanning/relevance_score.git}}.

Equation~\ref{eqn:localrel} allows us consider the descendants of a node independent of other branches arising from its ancestors. This can be applied recursively to calculate the local relevance score of the tree's root:~$\Xi(l, \underline{r}) = \Xi(l)$. 


The following statements are obviously true, and allow us to ignore the local relevance score $\Xi(l, \underline{n})$ for any nodes without descendants that have $label(\underline{n}) = l$:
\begin{align}
\shortintertext{ $\texttt{if } label(\underline{n}) = l $ }
\nonumber
    &\implies
        &\Xi(l, \underline{n}) =& 1
\shortintertext{ $ \texttt{if } T^{desc}_{\Pi}(\underline{n}) \cap L(l)  = \emptyset $ }
\nonumber
    &\implies
        &\Xi(l, \underline{n}) =& 0
\end{align}
Consider a node $ \underline{n} $ and one of its descendants \underline{d} such that all nodes with label $l$ that are a descendant of one are also a descendant of the other:
\begin{align}
\shortintertext{$     \texttt{if } T^{desc}_{\Pi}(\underline{n}) \cap L(l)  = T^{desc}_{\Pi}(\underline{d}) \cap L(l)  $ }
\nonumber
    &\implies
        &\Xi(l, \underline{n}) =&
            P\left( \underline{d} \in S_{\Pi}^{desc}(\underline{n}) \right) 
                &\times \Xi(l, \underline{d})
\\
\nonumber
        &&  =&
            P\left( \underline{d} \in S_{\Pi} | \underline{n} \in S_{\Pi} \right)
                &\times \Xi(l, \underline{d})
\\
\label{eqn:NodeToDesc}
        &&  =&
            \frac{\xi(\underline{d})}{\xi(\underline{n})}
                &\times \Xi(l, \underline{d})
\end{align}

\subsection{Derivation}
Now, consider Equation~\ref{eqn:XiFactBasic} when each child~$\underline{c}$ of a fact~node~$ \underline{f} $ has a single descendant~$ \underline{d} $ with~$label(\underline{d})~=~l$:
\begin{align}
\nonumber
    \Xi(l, \underline{f}) =&
        \sum_{ \underline{c} \in children(\underline{f}) }
            \frac
                { \Xi(l, \underline{c}) }
                { \left| children(\underline{f}) \right| }
\end{align}
Using Equation~\ref{eqn:CCAct}, this can be written as: 
\begin{align}
\nonumber
    =&
        \sum_{ \underline{c} \in children(\underline{f}) }
            \frac
                { \Xi(l, \underline{c}) } 
                {
                    \frac
                        { \xi(\underline{f}) }
                        { \xi(\underline{c}) }
                }
\\
\nonumber
    =&
        \sum_{\underline{c} \in children(\underline{f})}   
        \frac
            {\Xi(l, \underline{c}) \times \xi(\underline{c})} 
            {\xi(\underline{f})}
\end{align}
Next, using Equation~\ref{eqn:NodeToDesc}, this can be rewritten as:
\begin{align}
\nonumber
    =&
        \sum_{
            \substack
            {
                \underline{c} \in children(\underline{f})
                \\
                \underline{l}~\models T_{\Pi}^{desc}(\underline{c})~\cap~L(l) 
            }
        }
        \frac
            {
                \frac
                    { \xi(\underline{l}) }
                    { \xi(\underline{c}) }
                \times \Xi(l, \underline{l}) 
                \times \xi(\underline{c}) 
            } 
            {
                \xi(\underline{f})
            }
\\
\nonumber
    =&
        \sum_{ \underline{l}~\in~T_{\Pi}^{desc}(\underline{f})~\cap~L(l) }   
        \frac
            { \xi(\underline{l}) }
            { \xi(\underline{f}) }
\\
\label{eqn:XiFact}
    =&
        \frac
            { 1 }
            { \xi(\underline{f}) }
        \sum_{ \underline{l}~\in~T_{\Pi}^{desc}(\underline{f})~\cap~L(l) }
            \xi(\underline{l})
\end{align} 
If all descendants of a node $ \underline{d} \in  T_{\Pi}^{desc}(\underline{n}) $ are either such that $label(\underline{d})~=~l$, or are fact nodes for which Equation~\ref{eqn:XiFact} applies, this process can be repeated, causing further $ \frac{1}{\xi(\underline{c})} $ terms to cancel. If some nodes with $label(\underline{d_i})~=~l$ have an $ LCA(\underline{d_1},\underline{d_2})~=~\underline{a} $  that is an action (i.e., aLCA; see below), then Equation~\ref{eqn:NodeToDesc} does not apply between between $ \underline{c_i} $ and $ \underline{d_i}$. It will apply between $ \underline{a_i} $ and $ \underline{c_i} $, but $ \Xi(l,~\underline{a}) $  must be computed according to Equation~\ref{eqn:XiActBasic}:
\begin{align}
\nonumber
    \Xi(l, \underline{f}) =&
        \frac 
            { 1 } 
            { \xi(\underline{f}) }
        \sum_{ \underline{l} \in fLCAs }               
            \xi(\underline{l}) +
            \sum_{ \underline{a} \in aLCAs }
                \frac 
                    { \xi(\underline{a}) }
                    { \xi(\underline{f}) }
                \times
                \Xi(l, \underline{a})
\\
\label{eqn:XiAnyFact}
    =&
        \frac 
            { 1 } 
            { \xi(\underline{f}) }
        \left(
            \sum_{ \underline{l} \in fLCAs }
                \xi(\underline{l})
        \right.
        +
        \left.
        \sum_{ \underline{a} \in aLCAs }
            \xi(\underline{a}) \times \Xi(l, \underline{a})
        \right)
\end{align}
\filbreak
where:
\begin{itemize}
    \item aLCAs - \textbf{a}ction \textbf{L}owest \textbf{C}ommon \textbf{A}ncestors: any action nodes that are the LCA of any subset of $ L(l) \cap T_{\Pi}^{desc}(\underline{f})$
    \item fLCAs - \textbf{f}act \textbf{L}owest \textbf{C}ommon \textbf{A}ncestors: any nodes with label $ l $ whose paths diverge at fact nodes:
    \\
    $ 
    fLCAs =
        \left\{ 
            \forall \underline{l} \in L(l) :
                LCA(\underline{l}_i,\underline{l}_j) \in F \cap T_{\Pi}^{desc}(\underline{f})
        \right\}
    $ 
\end{itemize}
$\Xi(l)$ can be calculated from $L(l)$ by identifying the aLCAs, calculating $\Xi(l,\underline{c})$ for their direct children with Equation~\ref{eqn:XiAnyFact}, and combining them according to Equation~\ref{eqn:XiActBasic}. 

\subsection{Exploration}
For practical problems, $ T_{\Pi} $ is potentially very large, prohibiting its full representation.
A lower bound for $ \Xi(l) $ can be found using a partially explored tree $ \overline{T_{\Pi}} $. 
Nodes for which $ \xi(\underline{n}) $ is small contribute less to $ \Xi(l) $, and are found further from the root, causing this lower bound to converge quickly upwards as the region of the tree close to the root is explored. 

We use Algorithm~\ref{alg:exploration} to sample $ \overline{T_{\Pi}} $ from $ T_{\Pi} $.
The exploration threshold~$ \rho $ is an estimate of the ratio of how much information is present in the frontier compared to the explored tree. We found that the performance of $ h_{\Xi} $ as a heuristic is not sensitive to small changes in the value of $ \rho $. Unless stated otherwise, we use $ \rho = 0.2 $ in the remainder of this paper. The first term of the condition in the outer while loop in  Algorithm~\ref{alg:exploration} ensures that the sampling does not terminate early due to the volatility of its terms when small.

Note that this sampling procedure differs from the sampling performed by the hypothetical NDA. Both sampling procedures select actions that satisfy a frontier fact at random, but the NDA explores all facts that are preconditions for these actions, whereas the sampling procedure in Algorithm~\ref{alg:exploration} selects a single precondition. This ensures that the $ T_{\Pi} $ is sampled more diversely without too much focus on a few samples close to the root that branch widely, leading to their siblings being ignored.

\begin{algorithm}[tb]
\begin{algorithmic}[1]
\STATE Let $ frontier \leftarrow \{goal\}$ 
\STATE Let $ \overline{T_{\Pi}}  \leftarrow \{goal\}$ 
\WHILE{$ | \overline{T_{\Pi}} | < 100000$  OR \\$ sumxi(frontier) / sumxi(\overline{T_{\Pi}}) > \rho $ }
\label{algline:whileexpstart}
    \STATE Let $ \underline{n} \leftarrow choose (frontier) $ 
    \label{algline:choosedive}
    \WHILE{$ \underline{n} $ has children}
    \label{algline:choosechildrenstart}
        \STATE Remove $ \underline{n} $ from $ frontier $ 
        \STATE Add $ children( \underline{n} ) $ to $ frontier $ 
        \STATE Add $ children( \underline{n} ) $ to $ \overline{T_{\Pi}} $ 
        \STATE Let $ \underline{n} \leftarrow choose(children(\underline{n})) $ 
    \label{algline:choosechildrenend}
    \ENDWHILE
\label{algline:whileexpend}
\ENDWHILE
\end{algorithmic}
\hrulefill
\\
\textbf{function} $ sumxi(\texttt{K}):\\
\textbf{return} \sum_{\forall \underline{n} \in \texttt{K}} \xi(\underline{n})$\\
\hrulefill
\\
\textbf{function} $ choose(options): $ \\
\textbf{return} $\underline{m_i} \sim P(\underline{m_i}) = \xi(\underline{m})/\sum_{options}\xi(m_i) $ 
\caption{\textbf{Back-jumping depth-first tree search} 
Lines~\ref{algline:whileexpstart}~-~\ref{algline:whileexpend} start by exploring a minimum of 100000 nodes, and continue until the ratio of the sum of $\xi(\underline{n})$ in the frontier to that in the explored tree is less than $\rho$. 
Line~\ref{algline:choosedive} selects a node $\underline{n}$ from which to perform the next depth-first dive with probability proportional to $\xi(\underline{n})$.
Lines~\ref{algline:choosechildrenstart}~-~\ref{algline:choosechildrenend} perform a depth-first dive, adding the children of nodes explored to $ \overline{T_{\Pi}} $, but choosing one to explore at each depth. 
}
\label{alg:exploration}
\end{algorithm}

\subsection{Finding LCAs}
Much research has gone into finding the LCA of a pair of nodes in a tree. This problem has been translated into that of calculating Range Minimum Querys~\cite{Fischer2006}. Its complexity is $ O(h) $ to pre-process the tree, where $h$ is the height of the tree, and then $ O(1) $ for each query on a pair of nodes. This approach would thus cost $ O(h + n^2) $ to find LCAs for each pair of nodes, where $n$ is the number of nodes whose LCAs we want to find. 

Instead of this established procedure, we can take advantage of the fact that many LCAs will be shared by multiple pairs; there are at most $hn$ unique LCAs. We first sort the nodes by their paths (with complexity $O(hn\log(n)) $ ). The order does not matter, just that nodes with the same path up to a certain distance from the goal are in a contiguous group, so nodes are treated as symbols in a sequence (i.e., the path starting at the goal) and sorted alphabetically by their labels. Traversing along the sorted paths (with worst case complexity $ O(hn) $), checking for differences between nodes in paths (and whether or not they are action nodes) that are adjacent in the sort order yields an ordered list of action LCAs (aLCAs). Sets of aLCAs are computed once for each fact, and then filtered by $\sigma$ when $h_{\Xi}(\sigma)$ needs to be calculated.

\subsection{Use as a heuristic}
In any particular state ~$ \sigma $, all facts in the state $ f \in \sigma $ are true and do not need to be achieved by the planner. The state aware relevance score for a given label, $ \Xi_{\sigma}(l) $, should not consider parts of $ \overline{T_{\Pi}} $ that achieve these facts to be relevant. The relevance score applied to a state $ \Xi_{\sigma}(l) $ represents the probability that a node with label $ l $  would be sampled by the NDA, if it stops at nodes that are true in state $ \sigma $ (because they do not need an action to supply them). It is calculated according to Equations~\ref{eqn:XiAnyFact} and~\ref{eqn:XiActBasic} applied to the tree truncated at nodes with labels in $ \sigma $: 
\begin{equation}
\nonumber
    \overline{T_{\Pi}/\sigma} = \overline{T_{\Pi}} \left/ \bigcup_{\substack{
        \forall \underline{f} \in L(f) \\
        \forall f \in \sigma
    }} 
    T^{desc}_{\Pi}(\underline{f})
    \right.
\end{equation}
This is found for all facts, and the heuristic for a state is calculated as:
\begin{equation}
\nonumber
h_{\Xi}(\sigma) = \sum_{l \in F} \Xi_{\sigma}(l)
\end{equation}
As the relevance score heuristic is a sum of probabilities, $h_{\Xi}(\sigma) \in \mathbf{R}$. The LAMA planning system, on the other hand, requires heuristics to return an integer value. To allow for use and comparison with LAMA, our approach rounds $h_{\Xi}(\sigma)$ to the nearest integer, unfortunately losing some information in the process.

\section{Experimental Setup and Results}
We experimentally evaluated the following hypotheses:
\begin{itemize}
    \item[\textbf{H1}] The relevance score heuristic~\textbf{S1} is slower than the landmark counting heuristic~\textbf{S2} at solving standard planning problems, but is able to find a plan most of the time; and

    \item[\textbf{H2}] The relevance score heuristic~\textbf{S1} substantially improves the ability to find plans compared with the landmark counting heuristic~\textbf{S2} in domains without non-trivial landmarks. 
\end{itemize}
Hypothesis \textbf{H1} is based on the understanding that when the tree is fully explored (i.e. $ \overline{T_{\Pi}} = T_{\Pi} $), landmarks will have a relevance score $ \Xi(l) = 1 $ and facts that do not appear in any valid plans will have a relevance score $ \Xi(l) = 0 $. In other words, the use of the relevance score provides access to all the information available to a planning algorithm using the landmark counting heuristic, while also considering additional information about relevant but non-essential facts, which will result in additional computation. \textbf{H2} is valid because, while landmark counting heuristic~\textbf{S2} will have no non-trivial landmarks to guide its search, the relevance score~\textbf{S1} will still have an informative heuristic gradient to follow, with the potential to lead to (better) solutions to the planning problems. 


These hypotheses were evaluated using the architecture of the LAMA planner~\cite{Richter2010}\footnote{\label{foot:lama} \url{https://github.com/rock-planning/planning-lama.git}}, with either our relevance score heuristic~\textbf{S1} or the existing landmark counting heuristic~\textbf{S2}; the use of \textbf{S2} was our baseline. LAMA uses a weighted $A^*$ search procedure (using the default weight = 10) with one of the heuristics (\textbf{S1} or \textbf{S2}). All searches were subject to a RAM limit of 8GB. Their performance at solving all 674~problems defined in the examples folder of the \textbf{HSP2 repository}~\cite{Bonet2001}~\footnote{\label{foot:hsp2} \url{https://github.com/bonetblai/hsp-planners/tree/master/hsp2-1.0/examples}} was compared according to the following performance measures:
\begin{enumerate}
    \item[\textbf{M1}] Proportion of problems solved/not-solved;
    \item[\textbf{M2}] Time spent searching for a plan before finding one;
    \item[\textbf{M3}] Number of states expanded before finding a plan; and
    \item[\textbf{M4}] Length of plan found.
\end{enumerate}
These are well-established measures for evaluating the performance of these planning problems.

\subsection{Problems without non-trivial landmarks}
\label{sec:lmfgeneration}
In order to evaluate \textbf{H2}, we generated new PDDL specifications for domains that contain no landmarks other than facts in the goal and initial state. This was achieved by merging a pair of problems, $ \pi_1 $, $ \pi_2 $  that were consistently solved using both heuristics in such a way as to prevent them from interacting. This ensures that each new problem can be solved by at least 2 plans that have no overlap between the facts or actions involved in them.

To generate $ merged(\Pi_1, \Pi_2) $, all domain or problem specific elements (types, constants, predicates, actions, objects) in each problem were prepended with a label unique to that problem. This ensured that no element shared a name with elements in the problem they are being merged with. If a problem did not use types, a new type consisting of just its unique label was applied to all constants and objects within it. The new \textit{merged} problem is then comprised of all labeled elements from the two source problemss, with the following modifications:
\begin{itemize}
    \item Two new actions are defined:\\
        $ a_1: pre(a_1) = G_1; ef\!f(a_1) = \{winning\}$\\
        $ a_2: pre(a_2) = G_2; ef\!f(a_2) = \{winning\}$. 
    \item The merged problem has a single goal:\\ 
        $ G_{merged} = \{winning\}$ 
\end{itemize}
A total of 500 problems were generated with this procedure by randomly selecting a pair of problems from the pool of those solved individually by both \textbf{S1} and \textbf{S2}. The script used for this is unable to handle PDDL that, while valid, does not follow certain conventions. Whenever this occurred, a new pair was randomly selected for experimental evaluation.

\subsection{Results: Standard Problems}
\label{sec:results}
Of the 674 planning problems in the HSP2 examples repository, Table~\ref{tab:standardsolvefail} shows how many were solved using each heuristic (i.e., \textbf{S1}, \textbf{S2}) with the LAMA planner. Note that \textbf{S2} performed better but \textbf{S1} was able to find plans for most of the problems solved by \textbf{S2}.

\begin{table}[tb]
    \centering
    \begin{tabular}{c||c|c|c|c}
    Solved by               & \textbf{S1}   & \textbf{S2}   & Both      & Neither     \\
    \hline
    Number of               & 273           & 370           & 261       & 147         \\
    problems                & (51.61\%)     & (69.94\%)     & (49.34\%) & (27.79\%)              
    \end{tabular}
    \caption{Number of standard problems solved using \textbf{S1} and \textbf{S2} individually. \textbf{S2} performed better but \textbf{S1} was able to find plans most of the time.}
    \label{tab:standardsolvefail}
\end{table}

\begin{figure}[tb]
\begin{subfigure}{\linewidth}
    \includegraphics[width=\linewidth]{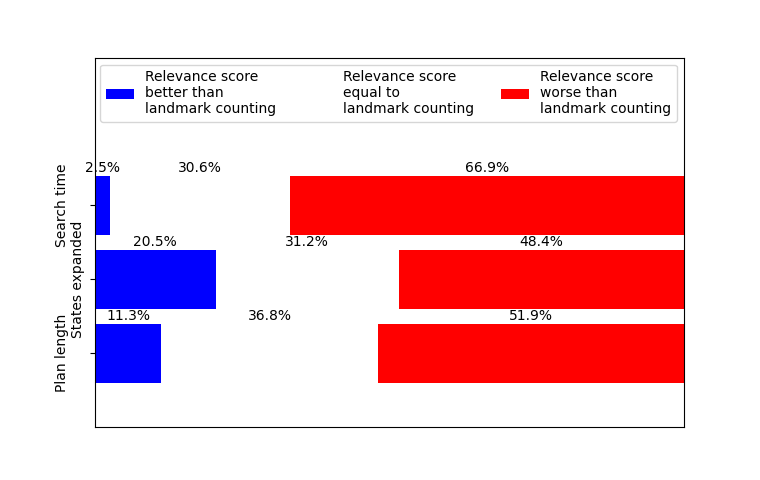}
\end{subfigure}
\caption{\textbf{Standard problems (674)}: for each measure, a lower value is better. \textbf{S2} (landmark-counting) is better than \textbf{S1} (relevance score) based on most measures; results support hypothesis \textbf{H1}.}
\label{fig:standardadversarial}
\end{figure}

Figure~\ref{fig:standardadversarial} shows that \textbf{S2} resulted in better performance than \textbf{S1} according to measures \textbf{M2}, \textbf{M3}, and \textbf{M4} for a majority of these problems. For problems that were solved by both heuristics:
\begin{itemize}
    \item the average ratio between search times (s) for $\textbf{S1}/\textbf{S2} = 951.34 \pm 3303.31$; 
    \item the average ratio between the number of states expanded for $\textbf{S1}/\textbf{S2} = 6.86 \pm 29.53$; and
    \item the average ratio between the length of plan found for $\textbf{S1}/\textbf{S2} = 1.13 \pm 0.29$.
\end{itemize}
These results were along expected lines and strongly support hypothesis \textbf{H1}.

\subsection{Results: Landmark-free problems}
Of the 500 problems generated according to the procedure outlined in Section~\ref{sec:lmfgeneration}, Table~\ref{tab:mergedsolvefail} summarizes how many were solved using each heuristic (i.e., \textbf{S1}, \textbf{S2}) with the LAMA planner. Note that performance was better using \textbf{S1} compared with \textbf{S2}.

\begin{table}[tb]
    \centering
    \begin{tabular}{c||c|c|c|c}
    Solved by               & \textbf{S1}   & \textbf{S2}   & Both      & Neither     \\
    \hline
    Number of               & 335           & 122           & 108       & 151            \\
    problems                & (67\%)        & (24.4\%)      & (21.6\%)  & (30.2\%)              
    \end{tabular}
    \caption{Number of merged problems solved using \textbf{S1} and \textbf{S2} individually. \textbf{S1} performed much better than \textbf{S2}, substantially improving the ability to solve the problems and find better plans.}
    \label{tab:mergedsolvefail}
\end{table}

\begin{figure}[tb]
\begin{subfigure}{\linewidth}
    \includegraphics[width=\linewidth]{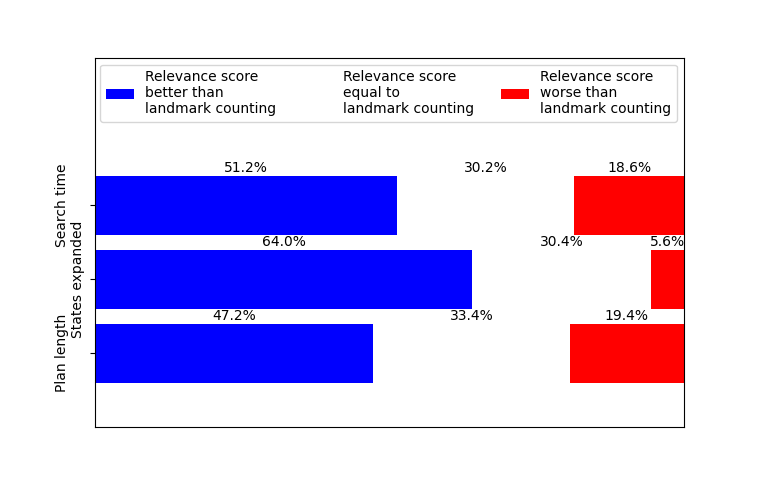}
\end{subfigure}
\caption{\textbf{Landmark free problems (500)}: for each measure, a lower value is better. \textbf{S1} (relevance score) is better than \textbf{S2} (landmark-counting) based on most measures; results support hypothesis \textbf{H2}.}
\label{fig:mergedadversarial}
\end{figure}

Figure~\ref{fig:mergedadversarial} shows that \textbf{S1} was measured to be much better than \textbf{S2} according to performance measures \textbf{M2}, \textbf{M3}, and \textbf{M4} for the majority of these merged problems. \textbf{S2} fails to solve most of this class of problems. For problems that were solved by both heuristics: 
\begin{itemize}
    \item the average ratio between search times (s) for $\textbf{S1}/\textbf{S2} = 123.08 \pm 295.25$;
    \item the average ratio between the number of states expanded for $\textbf{S1}/\textbf{S2} = 0.64 \pm 1.77$; and 
    \item the average ratio between the length of plan found for $\textbf{S1}/\textbf{S2} = 2.17 \pm 2.40$.
\end{itemize}
These results strongly support hypothesis \textbf{H2}.

\section{Discussion}
\label{sec:discuss}
The experimental results provide some interesting insights.
The relevance score calculated for a fully explored tree ($ \overline{T_{\Pi}} = T_{\Pi} $ ) is the number of facts that are landmarks for plans originating from $ \sigma $ (for which $ \Xi_\sigma(l) = 1$ ), added to the relevance calculated for other facts. 
In the case of standard planning problems, in addition to taking longer to compute and use this extra information (\textbf{M2}), the use of our proposed heuristic seems to impair the planners ability to find a plan within the resource limits imposed. Our explanation for this is that the relevant but not essential facts, for which $ \Xi_\sigma(l) $ is high but less than 1, guide the planner towards potentially competing plans. This "distraction" causes it to find plans that include elements of other partial plans that it could have found, leading to longer plans \textbf{M4}. In problems where there is less overlap between potential plans, the ratio of plan lengths (when both \textbf{S1} and \textbf{S2} find one) is greater. Plans found by \textbf{S1} in landmark-free problems contain actions from both source problems, whereas plans found by \textbf{S2} do not. The distracting information also increases the cost of exploration (\textbf{M2} and \textbf{M3}) which, causes more problems to be unsolvable within resource (RAM) limits. 

For landmark-free problems, \textbf{S2} can only tell that a partial plan might be good when it finds one of the goal facts. Until then, it searches a flat surface, increasing the distance from the initial state in all directions.  By contrast, \textbf{S1}, i.e., our proposed relevance score heuristic, is able to climb an informative surface that guides it toward potential plans. Because \textbf{S1} rewards the planner for finding facts that are relevant to alternative, but potentially separate plans, it has a tendency to include some actions in the final plan that did not contribute to achieving the goal. We believe this explains why it finds longer plans than \textbf{S2}, particularly on problems that are known to be solvable by disjoint plans. This is an unintentional byproduct of the way these domains were created that may not be the case on problems generated in other ways. 


Overall, the key observation is that our relevance score heuristic is able to guide the LAMA planner toward solving a class of planning problems for which landmark counting is ineffective. This comes at the cost of being more expensive to compute, and worse performance on standard problems. It can therefore only be recommended on the class of problems for which it is superior; those with few or no non-trivial landmarks. Note, however, that landmarks are identified before a plan search procedure begins,  which means that there is an easy and natural way to leverage both heuristics: \textit{choose landmark-counting on problems with well-defined landmarks, and use our relevance score heuristic for problems that only have trivial landmarks}. In future work, we hope to explore these insights further with additional (and more complex) planning problems.

\bibliography{library,GNT}

\end{document}